\documentclass[11pt]{article}

\usepackage[preprint]{acl}

\usepackage{times}
\usepackage{latexsym}
\usepackage[T1]{fontenc}
\usepackage[utf8]{inputenc}
\usepackage{microtype}
\usepackage{inconsolata}
\usepackage{graphicx}
\usepackage{amsmath,amssymb,amsfonts}
\usepackage{textcomp}
\usepackage{booktabs}
\usepackage{multirow}
\usepackage{array}
\usepackage{enumitem}
\expandafter\def\expandafter\UrlBreaks\expandafter{\UrlBreaks\do\-}
\title{Automated IEP Generation from Traditional Chinese Parent-Teacher Interviews\\
       via Corpus-Grounded Feature Diffusion}

\author{Kuanlin Chen \and Cheng-En Ou \\
  Department of Industrial Engineering and Engineering Management \\
  Chung Yuan Christian University, Taiwan \\
  \texttt{s11224246@cycu.edu.tw}}

\begin{document}
\maketitle

\begin{abstract}
Writing Individualized Education Programs (IEPs) is a heavy documentation burden for
special-education teachers, and privacy rules keep student records away from cloud AI
services. We ask whether a small language model that runs offline can match cloud
models at turning Traditional Chinese parent--teacher interviews into structured IEP
goals. We fine-tune the 7B-parameter Breeze model with QLoRA on 582 examples: 15
synthetic interviews with GPT-5.5 draft annotations reviewed and corrected by two
special-education teachers, and 567 further examples generated from them. On 10
held-out synthetic interviews with expert-reviewed annotations, the fine-tuned model is
closest to the experts in semantic similarity (BERTScore F1 0.779), ahead of four cloud
models used zero-shot: GPT-5.4 (0.726), DeepSeek-V3.2 (0.703), Gemini-3-Flash-Preview
(0.703), and Llama-4-Maverick (0.700); even the narrowest margin, over GPT-5.4, has a
paired bootstrap 95\% CI of 0.012 to 0.098. Without fine-tuning, the same model decoded
the same way scores lowest of all (0.653, against 0.752 after fine-tuning), so the gain
comes from the 582 examples. Fine-tuning also raises the share of parent quotes copied
verbatim from the interview from 25\% to 81\% (88\% in the best configuration). A
second finding runs against our own design. On 55 further synthetic interviews,
grammar-constrained decoding, added to guarantee valid output, lowered the share of
valid outputs from 55/55 to 51/55 for this model, because on four inputs it made the
output run past the token budget (exact McNemar p = 0.125, not significant at this
sample size). BERTScore measures similarity to expert annotations, not the educational
quality of the goals; expert rating of quality remains future work. Inference runs on a
single consumer GPU, so interview data need not leave the school.
\end{abstract}

\section{Introduction}

In modern special education systems, the drafting of Individualized Education Programs
(IEPs) and the management of parent--teacher interview records place an enormous burden
on educators. From an industrial engineering perspective, this represents a classic
\emph{high-labor, knowledge-intensive, repetitively format-constrained} information-processing
bottleneck. According to the description of an SBIR-funded IEP tool~\citep{itherapy2025},
special educators in the United States spend 6--10 hours drafting each IEP, with annual
reviews requiring a further 3--4 hours, a workload the award links to a 13.2\% attrition
rate and teacher burnout.

Large Language Models (LLMs) offer automation potential for such highly structured
professional text. In a study that randomly assigned 22 novice special-education
teachers to ChatGPT or control groups, \citet{rakap2024chatting} found that ChatGPT
assistance improved the rated quality of IEP goals for children with autism and reduced
the time spent writing them; \citet{rakap2024} report a preliminary study of
ChatGPT-assisted goal development for preschoolers with autism. Deploying generative AI
in special education nevertheless faces four challenges: (1)~domain data scarcity;
(2)~hallucination risk; (3)~cross-model evaluation bias; and (4)~Traditional Chinese
language adaptation.

Privacy adds a fifth constraint. A 2025 policy report by the Center for Democracy and
Technology (CDT)~\citep{cdt2025} cites a poll in which 57\% of teachers reported using AI
to develop an IEP or 504 plan in the 2024--25 school year, yet sending sensitive
disability-related data to cloud APIs risks violating the U.S.\ Family Educational
Rights and Privacy Act (FERPA) and the Individuals with Disabilities Education Act
(IDEA). We therefore pursue a \emph{local fine-tuning} strategy based on an open-source
small model. Our goal is to transform Traditional Chinese parent--teacher interview
transcripts into a hierarchical SMART Goal Ladder that complies with Article~31 of
Taiwan's Special Education Act and the 114th Academic Year Preschool IEP Form, with
schema-constrained JSON output.

\textbf{Contributions.} This work makes four contributions to the under-resourced field
of Traditional Chinese special-education NLP:

\begin{enumerate}[leftmargin=*]
  \item \textbf{Methodological transfer:} We extend the corpus-grounded synthetic
        methodology from industrial maintenance work orders~\citep{lau2025} to educational
        service systems as \emph{Corpus-Grounded Feature Diffusion} (CGFD), a six-stage
        pipeline: seed overproduction $\rightarrow$ $\tau$ threshold selection
        $\rightarrow$ dual-expert annotation $\rightarrow$ FeatureProfile extraction
        $\rightarrow$ diversity diffusion $\rightarrow$ QLoRA fine-tuning and local
        inference.
  \item \textbf{Anti-mode-collapse design:} An \emph{archetype} $\times$
        \emph{speaker} $\times$ \emph{rationalization} three-layer quota
        ($8 \times 5 \times 12$) combined with \emph{six-layer detection}
        (prompt-leak / model-artifact / repeated parent opening /
        repeated teacher opening / oversampled $n$-gram / pairwise Jaccard)
        and hard/soft regeneration, designed to counter stylistic
        collapse~\citep{vs2025}; the 80 seed candidates reach Distinct-4 = 0.787
        and Self-BLEU = 0.618.
  \item \textbf{Expert-adjudicated AI pre-annotation:} An
        \emph{LLM-assisted pre-annotation with expert adjudication} workflow
        in which GPT-5.5 drafts IEP fields and experts review and correct them,
        turning blank-page labeling into field-level adjudication; acceptance rate,
        correction rate, and time reduction quantify the annotation burden.
  \item \textbf{Cross-vendor evaluation design:} A staggered vendor design
        (GPT-5.4 for data generation, Kimi-K2.6 for pre-screening, and GPT-5.4,
        DeepSeek-V3.2, Gemini-3-Flash-Preview, and Llama-4-Maverick as zero-shot
        baselines; Table~\ref{tab:vendors}) evaluated against expert gold annotations,
        plus a specified \emph{centered}-DBG protocol~\citep{chen2025dbg} for auditing
        LLM-as-Judge self-preference bias~\citep{wataoka2024}, whose run is left to
        future work.
\end{enumerate}

\textbf{Findings.} On a 10-interview hold-out set, the fine-tuned Breeze-7B without
grammar-constrained decoding (GCD) is the closest of all tested systems to the expert
annotations in BERTScore F1 (0.779). It is ahead of all four zero-shot cloud baselines,
and even its narrowest margin, over GPT-5.4, has a paired bootstrap 95\% CI of $+0.012$
to $+0.098$. The gain comes from fine-tuning rather than decoding: under identical GCD
decoding, fine-tuning raises F1 from 0.653 to 0.752 and the share of verbatim parent
quotes from 25\% to 81\%. GCD itself, added as a reliability guardrail, did not help
this adapter: on a 55-interview stress set it produced the only four invalid outputs,
because it lengthened the output past the token budget; follow-up runs suggest that
the 2048-token training cap and the decoding pipeline both play a role. All inference
runs on a single
consumer GPU, so interview data need not leave the school.

\section{Related Work}

\subsection{Seed-Based Synthetic Data Generation}
The paradigm of amplifying training data from a small seed set in low-resource settings was
established by Self-Instruct~\citep{wang2022selfinstruct}, which iteratively generated tens of
thousands of instruction instances from 175 human-authored seed tasks via LLM self-iteration.
Stanford Alpaca~\citep{taori2023alpaca} applied the same concept to fine-tune LLaMA-7B on 52K
instruction pairs generated by \textsc{text-davinci-003}, establishing the industrial
benchmark of ``closed-source large model generates, open-source small model absorbs.''
WizardLM's Evol-Instruct~\citep{xu2023wizardlm} introduced depth and breadth evolution
mechanisms to address instruction-complexity stagnation. More recently, Auto
Evol-Instruct~\citep{zeng2024autoevol} enables LLMs to automatically analyze data
characteristics and devise evolution strategies. The present work extends this paradigm
to structured annotation diffusion in the education domain.

\subsection{Corpus-Grounded Synthetic Data}
Unconstrained diffusion tends to produce homogeneous synthetic text with degraded domain
authenticity. \citet{lau2025} generate synthetic maintenance work orders grounded
in real engineering terminology and technician language patterns extracted from the MaintIE
annotation set~\citep{bikaun2024maintie}. Similarly, SQaLe~\citep{wolff2025sqale} uses real
database schemas as Text-to-SQL synthesis constraints, while SPICE~\citep{liu2025spice}
employs corpus-environment self-play to improve reasoning generation. The authors' prior
\textsc{ARID} system~\citep{authors2026arid} extends this methodology as
\emph{Noise-Aware Synthetic Distillation (NASD)}, building a noise-aware pipeline
for industrial maintenance work orders deployable on edge hardware in an air-gapped
setting. The present study \emph{transfers} the same methodology to Traditional
Chinese special education, extracting a FeatureProfile (sentence length, structure,
quantification templates) from the 15 expert-annotated synthetic training seeds as
diffusion constraints, with dual-expert annotation and $\tau$ selection as entry criteria.

\subsection{LLM-as-Judge Self-Preference Bias}
Using large closed-source models as automatic judges for fine-tuned open-source models is
standard practice, yet research reveals a systematic Self-Preference Bias (SPB).
\citet{wataoka2024} show that GPT-4 prefers text with lower perplexity,
i.e., text more similar to its own output distribution. The DBG Score of
\citet{chen2025dbg} introduces a gold standard as a quality proxy to decouple
generation capability from preference bias; \citet{yang2026spb} further propose
an automated framework that reduces SPB by an average of 31.5\% without human intervention.
InstaJudge~\citep{jang2025instajudge} provides an industry-grade distribution-preserving
few-shot sampling solution. This work adopts a staggered multi-vendor design
(Table~\ref{tab:vendors}) evaluated against expert gold annotations, and specifies
\emph{centered}-DBG --- subtracting $J$'s global scoring tendency toward other models
from $\text{DBG}_J(J)$ --- as a protocol for separating self-preference from
general-strictness bias without relying on any single judge; running it is future
work (Section~\ref{sec:spb}).

\subsection{Grammar-Constrained Decoding (GCD)}
\citet{geng2023gcd} showed that GCD can strictly restrict token generation to a
context-free grammar (CFG) at inference time, achieving structured output without additional
fine-tuning. \citet{schmidt2025gcdie} show that GCD raises F1 for fine-tuned
encoder--decoder models in low-resource information extraction from clinical trial
abstracts. Open-source frameworks such as \textsc{Outlines}, \textsc{Guidance},
and \textsc{XGrammar}~\citep{dong2025xgrammar} precompile JSON Schema via pushdown-automata
(PDA) technology, enabling microsecond-level token-mask computation~\citep{saibo2024gcd}.
We use \textsc{Outlines} v1.2 with Pydantic models; for our adapter, however, GCD did
not improve reliability under a fixed 2048-token budget (Sections~\ref{sec:ablation}
and~\ref{sec:discussion}).

\subsection{IEP NLP Automation}
In English-language research, the randomized study of \citet{rakap2024chatting} shows
that ChatGPT assistance improves the quality of IEP goals written by novice
special-education teachers and reduces writing time, and \citet{rakap2024} extend this to
preschoolers with autism in a preliminary study. \citet{waterfield2025} examine ethical
safeguards in teacher training, and the CDT report~\citep{cdt2025} documents widespread
AI use for IEPs alongside FERPA/IDEA privacy risks. Commercially,
iTherapy~\citep{itherapy2025} received
NCSER SBIR funding to develop an \emph{Automated Compliance and Communication System for
IEP Management} targeting a 30\% reduction in drafting time. In Traditional Chinese,
relevant NLP applications are extremely sparse; the CDITS dual-agent dialogue tutoring system
of \citet{liao2025cdits} is among the few empirical examples, while research
directly addressing Traditional Chinese IEP automation remains absent.

\subsection{Traditional Chinese Foundation Model Adaptation}
\textsc{Breeze-7B}~\citep{hsu2024breeze7b} expands the original Mistral-7B vocabulary from
32K to 62K tokens, doubling Traditional Chinese compression efficiency and inference speed,
with 32K long-context support. The subsequent Breeze~2~\citep{hsu2025breeze2} further
integrates multimodal and function-calling capabilities. This system selects Breeze-7B as
the fine-tuning backbone so that the model focuses on aligning IEP domain semantics rather
than re-learning Chinese grammatical structure from scratch.

\subsection{Mode Collapse and Diversity Control}
LLMs tend to generate ``typical, safe, but homogeneous'' text following SFT and
RLHF~\citep{vs2025}. Verbalized Sampling~\citep{vs2025} advocates prompt-based strategies ---
instructing the model to jointly produce multiple candidates with probability distributions ---
to relieve typicality pressure. N-gram style priors~\citep{ngram2026} intervene directly in
logit space at decoding time. \citet{wright2026epistemic} measure epistemic diversity, the
diversity of real-world claims in model outputs, beyond lexical diversity. This work
controls style distribution via an
archetype $\times$ speaker $\times$ rationalization three-layer quota
($8 \times 5 \times 12 = 480$ combinations) combined with 7-gram repetition detection
that automatically triggers regeneration.

\subsection{Inter-Rater Reliability}
\citet{james2026iaa} notes that as annotation tasks scale from simple classification to
semantic mapping, Cohen's weighted $\kappa$ (two annotators), Fleiss's $\kappa$ (multiple
annotators), and Krippendorff's $\alpha$ (imbalanced data) each have appropriate use cases.
The LEAP protocol of \citet{lee2023common} shows that raising agreement through tailored
guidelines can make annotations diverge across research groups, so reliability is not a
proxy for validity. The MoA framework~\citep{moa2026} proposes multi-agent
collaborative annotation to address human-resource bottlenecks in low-resource settings.
This system computes both Cohen's weighted $\kappa$ and Krippendorff $\alpha_{\text{interval}}$
during seed construction, treating them as descriptive statistics for rating-procedure
consistency rather than standalone quality gates. Seed selection combines the dual-expert
high-score intersection, auto-flag hard/soft gates, and pairwise-overlap limits;
Section~\ref{sec:phaseB} reports how far the compressed ratings actually determined the
selected set. Boundary cases in annotation are retained in a manual adjudication queue
following LEAP~\citep{lee2023common}.

\section{Methodology}

\subsection{System Architecture}
\begin{figure*}[!t]
\centering
\includegraphics[width=\textwidth]{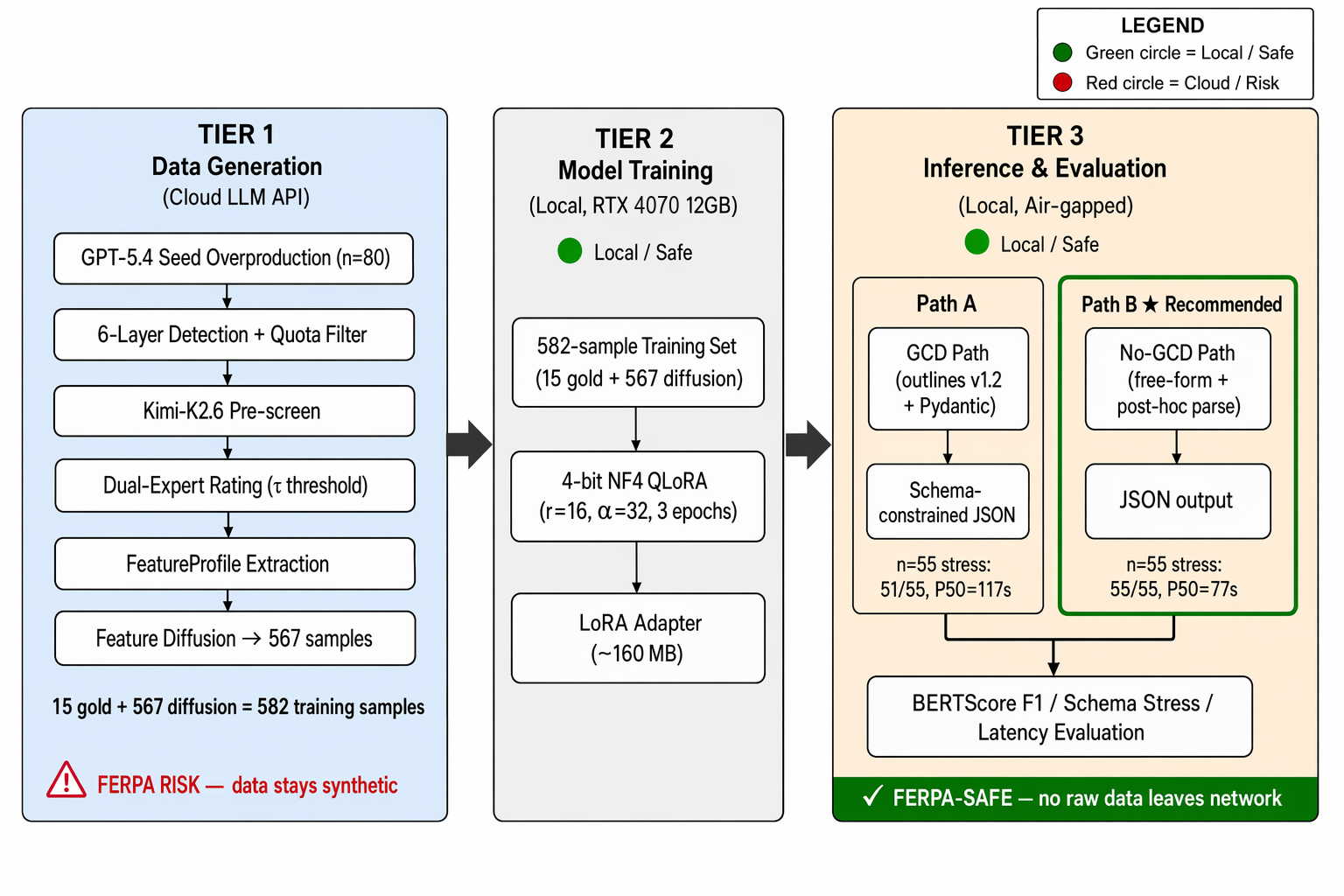}
\caption{Three-tier system architecture: data generation (Cloud LLM API) /
training (RTX 4070) / inference (local, FERPA-safe).}
\label{fig:arch}
\end{figure*}

The system consists of three tiers, as illustrated in Fig.~\ref{fig:arch}:
the \emph{data generation tier} (cloud LLM API) handles seed overproduction and feature
diffusion; the \emph{model training tier} (single RTX 4070, 12 GB VRAM) executes
4-bit QLoRA fine-tuning; and the \emph{inference and evaluation tier} performs local
offline inference (with and without GCD) and multi-axis evaluation.

\subsection{Schema Design: Hierarchical SMART Goals}
To align with Article 31 of Taiwan's Special Education Act and the 114th Academic Year
Preschool IEP Form, we design a three-layer Pydantic schema:
\begin{itemize}[leftmargin=*]
  \item \textbf{ParentConcern} --- concerns extracted from the interview, containing
        \texttt{raw\_quote}, \texttt{domain} (one of six developmental domains),
        and \texttt{is\_primary} (exactly one entry must be \texttt{True}).
  \item \textbf{SmartGoal} --- a five-element SMART-structured goal; \texttt{behavior}
        must be positively framed (no negation prefixes such as ``do not'' or ``avoid''),
        and \texttt{mastery} must be quantifiable (count/minute/percentage).
  \item \textbf{GoalLadder} --- one \texttt{annual\_goal} (high-level direction) paired
        with 2--3 \texttt{semester\_goals} (concrete step instances).
\end{itemize}
The six developmental domains align with Taiwan's early-intervention curriculum:
Cognitive, Communication, Motor, Social-Emotional, Sensory Processing, and Self-Care.

\subsection{Phase A: Seed Overproduction, Pre-screening, and Rating Preparation}
\subsubsection{Diversity Quota Assignment}
\label{sec:quota}
To prevent stylistic mode collapse~\citep{vs2025}, each candidate sample is assigned
metadata drawn from three independent pools:

\begin{itemize}[leftmargin=*]
  \item \textbf{Archetype (parent tone prototype, 8 types):}
        Anxious-compliant, Defensive-denial, Help-seeking-confused,
        Over-compliant, Professional-skeptic, Peer-comparing,
        Acceptance-planning, Plain-worrier.
  \item \textbf{Speaker (primary narrator, 5 types; weighted by Taiwan family structure):}
        Mother 0.50 / Maternal grandmother 0.20 / Father 0.15 /
        Other primary caregiver 0.10 / Paternal grandfather 0.05.
  \item \textbf{Rationalization (justification pattern, 12 types, 25\% None):}
        Gender stereotype, generational gap, family genetics, positive reframe,
        peer comparison, developmental delay attribution, parenting conflict,
        environmental attribution, health attribution, etc.
\end{itemize}

\subsubsection{Six-Layer Detection and Dual-Layer Regeneration}
To combat multiple LLM degradation modes in batch synthesis, post-generation
\emph{six-layer automatic detection} is applied, classified into hard/soft tiers by severity:

\begin{itemize}[leftmargin=*]
  \item \textbf{Hard reject} (mandatory regeneration):
        \emph{prompt-leak} (literal prompt example residue),
        \emph{model-artifact} (LLM refusal remnants),
        \emph{repeated parent opening} (first 10 characters identical),
        \emph{repeated teacher opening} (first 10 characters identical).
  \item \textbf{Soft regen} (at most 30\% of the batch regenerated per round):
        \emph{oversampled $n$-gram} (occurrence rate $>10\%$),
        \emph{pairwise $7$-gram Jaccard} $\geq 0.30$.
\end{itemize}

Upon regeneration, all pragmatic conditions (archetype, speaker, rationalization,
family\_config) are resampled so that the new sample does not inherit the old one's
stylistic inertia, and a ``banned phrase list'' is injected into the prompt as negative
feedback. The speaker is redrawn uniformly from the other four types, which flattens the
intended speaker weights (Section~\ref{sec:phaseA}).
Each sample also carries a \emph{per-sample retry} mechanism: if the initial generation
triggers a prompt-leak or artifact flag, up to 3 automatic retries are attempted before
the sample is marked as failed. The complete quality gate is:
up to 3 batch-level regeneration rounds $\times$ 3 per-sample retries.

\subsection{Phase B: $\tau$ Selection, Flag-aware Score Caps, and Annotation Package}
\label{sec:phaseB-method}
Two special-education teachers independently rated each of the 80 candidates on a
1--5 \emph{transcript realism} scale (1 = clearly machine-generated; 5 = reads like
a real recording, with natural fillers, self-correction, and affect). The scale
measures how authentic a transcript sounds, not the quality of any IEP derived from it.
Samples flagged \texttt{prompt\_leak} or \texttt{model\_artifact} are excluded before
ranking. To keep samples with high expert scores but residual auto-detected anomalies
out of the training set, a \emph{flag-aware cap} is applied to each rater's score
$s_r$, $r \in \{A, B\}$:
\begin{equation*}
\text{eff}_r = \min\!\big(s_r,\;
                     \min_{f \in \text{flags}} c_f \big)
\end{equation*}
where $c_f$ denotes the score ceiling for each flag:
$c_{\text{model\_artifact}}=1$,
$c_{\text{prompt\_leak}}=2$,
$c_{\text{repeated\_opener}}=3$,
$c_{\text{high\_ngram\_overlap}}=3$,
$c_{\text{repeated\_teacher\_opener}}=4$,
$c_{\text{phrase\_collision}}=4$.
The first two ceilings apply only when hard exclusion is disabled; with the default
configuration such samples never reach ranking. The hard-reject flags of Phase~A trigger
regeneration, but a sample that still carries a flag after the final round stays in the
candidate pool with that flag, so the remaining caps do apply.

\begin{figure}[!t]
\centering
\includegraphics[width=\columnwidth]{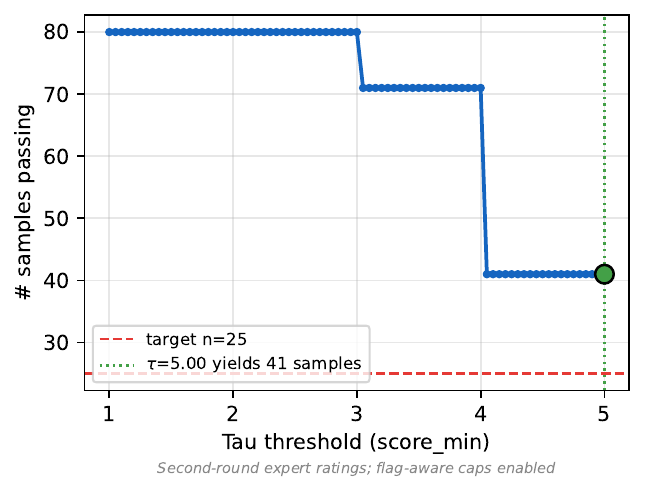}
\caption{Number of candidates passing each $\tau$ threshold on \texttt{score\_min}
(Phase B selection). Red dashed line: target of 25 seeds; green dotted line: adopted
$\tau = 5$, which 41 candidates pass. Computed from the second-round expert ratings with
flag-aware score caps, as a record of the selection decision.}
\label{fig:tau}
\end{figure}

Samples are ranked by $\texttt{score\_min} = \min(\text{eff}_A, \text{eff}_B)$,
with ties broken by \texttt{score\_avg}, the mean of the two effective scores.
Selection then proceeds greedily down the ranking under three diversity constraints (at most five samples per archetype, one
per parent-opening prefix, and pairwise character 7-gram Jaccard below 0.30) until
25 samples are chosen; these are split at random (seed 42) into 15 training and
10 hold-out seeds. Candidates still tied after both keys keep their generation order.
The minimum \texttt{score\_min} among selected samples is recorded as the operative
$\tau$ threshold and plotted against pass rate for traceability (Fig.~\ref{fig:tau}).

Rating agreement is reported as exact and adjacent (within one point) agreement,
Cohen's weighted $\kappa$, and Krippendorff's $\alpha_{\text{interval}}$~\citep{james2026iaa}.
Because $\kappa$-type coefficients become unstable when ratings concentrate in one
category~\citep{feinstein1990paradox}, we also report Gwet's AC1 and AC2, which are
designed to stay stable under such concentration~\citep{gwet2008ac1}.
To test how much the ratings, rather than tie order, determined the selected set, we
re-run the selection under alternative rules: each rater alone, mean-first ranking,
no flag caps, reversed tie order, and 1{,}000 random draws each with random
tie-breaking, with ratings ignored among flag-free candidates, and with ratings
ignored entirely.

\subsection{Phase C0: AI-Assisted Pre-annotation and Expert Adjudication}
Motivated by evidence that ChatGPT assistance improves IEP goal quality and reduces
writing time~\citep{rakap2024chatting}, an \emph{AI pre-annotation} subprocess is inserted
between Phase B and Phase C to measure, in a Traditional Chinese context, how much AI
drafts reduce expert effort:
\textsc{GPT-5.5} generates weak pre-annotations for 25 seed transcripts, populating
the \texttt{answer} column of a CSV template. Experts revise, delete, or rewrite
individual fields rather than filling from blank; the final gold label is always the
expert-corrected version, and AI drafts never enter the training set directly.

\noindent
\textbf{Three quantitative metrics:}
\begin{itemize}[leftmargin=*]
  \item \emph{acceptance\_rate}: fraction of AI draft fields accepted verbatim by experts.
  \item \emph{correction\_rate}: fraction of draft fields modified by experts.
  \item \emph{time\_reduction\_rate}: percentage decrease in AI-assisted annotation time
        relative to a blank-labeling time estimate; coverage is reported as
        \emph{time\_coverage\_rate}.
\end{itemize}
This design simultaneously provides (a) RCT-style engineering evidence and
(b) a quantitative time-ROI estimate for weak-supervised generation in low-resource domains.
Critically, GPT-5.5 drafts are \textbf{not treated as gold labels}; only expert-adjudicated,
Phase-C-validated versions enter the training set.

\subsection{Phase C: FeatureProfile Extraction and Diffusion Generation}
\label{sec:phaseC-method}
Phase C activates only after both experts' final CSV files are collected.
The system first parses both expert annotations as JSON, computes field-level agreement
rates (\texttt{primary\_domain}, \texttt{concern\_count}, \texttt{has\_secondary}),
and issues three classes of validation warnings: (i)~\emph{negative-prefixed behavior};
(ii)~\emph{is\_primary count} $\neq 1$; (iii)~\emph{semester\_goals count} $\notin [2,3]$.
Samples with any dual-expert field disagreement or validation warning are written to an
\texttt{adjudication\_queue.csv} and \textbf{excluded} from \texttt{parsed\_gold.json}
by default, per the LEAP auditable-adjudication requirement~\citep{lee2023common}.

From the 15 training-split seeds, the following distributional features $\mathcal{F}$
are computed --- each retaining mean / stdev / min / max / $p_{50}$ five-point statistics:
{\footnotesize
\begin{align*}
\mathcal{F}_{\text{length}}
  &= \{\texttt{transcript\_char\_len},\;
       \texttt{turn\_count},\\[-3pt]
  &\quad\;\texttt{annual\_behavior\_len},\\[-3pt]
  &\quad\;\texttt{semester\_behavior\_len},\\[-3pt]
  &\quad\;\texttt{context\_len},\;
       \texttt{mastery\_len}\}\\[2pt]
\mathcal{F}_{\text{structure}}
  &= \{\texttt{concerns\_per\_sample},\\[-3pt]
  &\quad\;\texttt{smart\_goals\_per\_sample},\\[-3pt]
  &\quad\;\texttt{semesters\_per\_ladder},\;
           \texttt{secondary\_ratio}\}\\[2pt]
\mathcal{F}_{\text{lexical}}
  &= \{\texttt{mastery\_patterns},\;
       \texttt{domain\_freq}\}
\end{align*}
}
\texttt{mastery\_patterns} samples the 30 most frequent quantification templates from
the expert gold annotations (e.g., ``$X$ out of $Y$ trials per week,'' ``no more than $Z$ occurrences'')
as literal corpus anchors for diffusion.

Each diffusion prompt attaches 5 randomly sampled expert-annotated training seeds as in-context
exemplars, combined with the $\mathcal{F}$ summary and schema rules.
The production run targets 585 samples; 567 valid diffusion samples pass schema and
quality gates (18 invalid/error); the final training set is 15 expert gold + 567
diffusion samples = 582 total. Domain quota aligns with Taiwan early-intervention
literature: Communication 30\% / Cognitive 22\% / Social-Emotional 22\% /
Motor 14\% / Self-Care 8\% / Sensory 4\%.

\subsection{Phase D: QLoRA Fine-Tuning}
\label{sec:phaseD}
We apply 4-bit NF4 double quantization (bitsandbytes) with LoRA $r=16$, $\alpha=32$,
dropout=0.05, targeting modules \texttt{q/k/v/o\_proj} and
\texttt{gate/up/down\_proj}. Training runs for 3 epochs, per-device batch size 1,
gradient accumulation 8 (effective batch 8), $\eta = 2 \times 10^{-4}$ with a cosine
schedule and 5\% warmup, a maximum sequence length of 2048 tokens, and gradient
checkpointing. The 582 examples are split 90/10 (seed 42) into 524 training and 58
validation examples. The 2048-token training cap is tight for this data: 492 of the 582
examples (444 of the 524 training examples) are longer, so the final part of their target
outputs is cut off, about 98k target tokens in total
(\texttt{20\_check\_training\_truncation.py}). The LoRA adapter is output to
\texttt{models/iep\_v3\_qlora}; the resulting
\texttt{adapter\_model.safetensors} is approximately 160 MB.

\subsection{Phase E: Local Inference}
On the GCD path, \textsc{Outlines} v1.2 compiles the Pydantic model into
a token-level mask~\citep{geng2023gcd,dong2025xgrammar}, restricting decoding to the
schema-valid token space (Grammar-Constrained Decoding, GCD). A no-GCD path
(free-form generation + post-hoc JSON parsing) is evaluated in parallel to avoid
conflating method-layer guarantees with empirical necessity; it produced no invalid
outputs on the stress set (Section~\ref{sec:ablation}). The Mistral-style prompt
template is:
{\footnotesize\begin{verbatim}
<s>[INST] {SYS_PROMPT}
[transcript] {transcript_text} [/INST]
\end{verbatim}}
Decoding is greedy, with an inference budget of \texttt{max\_new\_tokens} = 2048. This
budget is separate from the 2048-token training cap of Phase~D;
Section~\ref{sec:ablation} analyzes how the two relate to the GCD failures.

\subsection{Cross-Vendor Evaluation Framework}
To mitigate self-preference bias~\citep{wataoka2024,chen2025dbg}, the evaluation
assigns pipeline roles to different vendors where possible (Table~\ref{tab:vendors}).

\begin{table}[!t]
\caption{Cross-Vendor Role Assignment}
\centering
\footnotesize
\setlength{\tabcolsep}{4pt}
\begin{tabular}{@{}lll@{}}
\toprule
\textbf{Role} & \textbf{Model} & \textbf{Vendor} \\
\midrule
Data generation     & GPT-5.4                & OpenAI \\
Pre-screening       & Kimi-K2.6              & Moonshot \\
Gold pre-annotation & GPT-5.5                & OpenAI \\
Zero-shot baselines & GPT-5.4                & OpenAI \\
                    & DeepSeek-V3.2          & DeepSeek \\
                    & Gemini-3-Flash-Preview & Google \\
                    & Llama-4-Maverick       & Meta \\
System under test   & Breeze-7B + QLoRA      & MediaTek$^\ast$ \\
Gold standard       & Two experts            & --- \\
\midrule
\multicolumn{3}{@{}l}{\scriptsize $^\ast$Base model; fine-tuned locally by the authors.} \\
\bottomrule
\end{tabular}
\label{tab:vendors}
\end{table}

\subsection{Phase F: Paper-Level Quantitative Metrics}
All metrics are centralized in \texttt{\_metrics.py}
(\texttt{distinct\_n}, \texttt{self\_bleu}, \texttt{krippendorff\_alpha},
\texttt{dbg\_summary}), exposed via \texttt{08\_compute\_paper\_metrics.py}
and \texttt{09\_compute\_dbg\_score\_spb.py}.

\begin{itemize}[leftmargin=*]
  \item \emph{Lexical diversity} (character-level, normalized, punctuation/whitespace
        removed): Distinct-1 through Distinct-4~\citep{li2016diversity}, and
        Self-BLEU (4-gram, add-one smoothing)~\citep{zhu2018texygen}.
  \item \emph{Categorical entropy}: Shannon entropy across archetype, speaker,
        and rationalization dimensions to detect batch-level style collapse.
  \item \emph{Annotation reliability}: Cohen's weighted $\kappa$ and
        Krippendorff $\alpha_{\text{interval}}$~\citep{james2026iaa};
        the latter is computed via coincidence matrix and is particularly suited
        to imbalanced data with continuous scoring.
  \item \emph{Uncertainty}: paired percentile bootstrap (10{,}000 resamples) and exact
        sign-flip permutation tests over hold-out records
        (\texttt{19\_bertscore\_bootstrap.py}).
  \item \emph{Self-Preference Bias} (protocol only, not yet run): $\text{DBG}_J(J)$
        and $\text{centered-DBG}_J$ per Eqs.~(\ref{eq:dbg})--(\ref{eq:cdbg}).
\end{itemize}

The DBG-Score input contract (CSV fields: \texttt{sample\_id},
\texttt{judge\_model}, \texttt{response\_model}, \texttt{judge\_score},
\texttt{gold\_score}, \texttt{is\_self}, \texttt{notes}) is auto-generated
via \texttt{--make\_template}. Full SPB quantification awaits cross-model
judge/gold score collection in future work.

\section{Experiments}

\subsection{Phase A: Seed Generation Quality}
\label{sec:phaseA}
GPT-5.4 overproduced 80 seed candidate transcripts (quality gate configured
for up to 3 batch-level regeneration rounds and 3 per-sample retries), supplemented by
Kimi-K2.6 independent pre-screening. Table~\ref{tab:phaseA-stats} summarizes
the diversity metrics of the final batch ($n=80$). Archetype entropy is 2.957~bits
(uniform maximum $\log_2 8 = 3.000$) and rationalization entropy 3.209~bits (uniform
maximum $\log_2 12 = 3.585$), so these two pools spread samples across their
categories. Speaker entropy is 2.293~bits, close to the uniform maximum
$\log_2 5 = 2.322$ and above the 1.92~bits implied by the intended speaker weights
(Section~\ref{sec:quota}); each of the five speaker types occurs 11--20 times, because
regeneration redraws the speaker uniformly.

\begin{table}[!t]
\caption{Phase A Seed Generation Diversity Metrics ($n=80$).
$\uparrow$/$\downarrow$: higher/lower means more diverse.}
\centering
\footnotesize
\begin{tabular}{@{}lrr@{}}
\toprule
\textbf{Metric} & \textbf{Value} & \textbf{Uniform max} \\
\midrule
$n$ samples                  & 80           & --- \\
avg.\ characters / sample    & 1{,}789.5    & --- \\
\midrule
\multicolumn{3}{l}{\emph{Categorical entropy}} \\
archetype entropy            & 2.957        & 3.000 \\
speaker entropy              & 2.293        & 2.322 \\
rationalization entropy      & 3.209        & 3.585 \\
\midrule
\multicolumn{3}{l}{\emph{Lexical diversity (character-level)}} \\
Distinct-1 ($\uparrow$)      & 0.0108       & --- \\
Distinct-2 ($\uparrow$)      & 0.2207       & --- \\
Distinct-3 ($\uparrow$)      & 0.5495       & --- \\
Distinct-4 ($\uparrow$)      & 0.7870       & --- \\
Self-BLEU, 4-gram ($\downarrow$) & 0.6175   & --- \\
\bottomrule
\end{tabular}
\label{tab:phaseA-stats}
\end{table}

Kimi-K2.6 passed all 80/80 samples in the final batch (1--5 realism scale; score 5: 65;
score 4: 15).
This batch was then passed through six-layer automatic detection
(\emph{prompt-leak} / \emph{model-artifact} / \emph{repeated parent opening} /
\emph{repeated teacher opening} / \emph{oversampled $n$-gram} / \emph{pairwise Jaccard
$\geq 0.30$}) before submission to dual-expert scoring.
Each sample's \texttt{auto\_flags} (seven categories: \texttt{ok},
\texttt{model\_artifact}, \texttt{prompt\_leak}, \texttt{repeated\_opener},
\texttt{repeated\_teacher\_opener}, \texttt{high\_ngram\_overlap},
\texttt{phrase\_collision}) were recorded in the scoring CSV for expert reference.
Self-BLEU 0.6175 and Distinct-4 0.7870 serve as reference values for the diffusion
batches in Table~\ref{tab:ablation}.

\subsection{Phase B: $\tau$ Threshold Selection}
\label{sec:phaseB}
\paragraph{Rating distribution and agreement}
Realism ratings were concentrated at the top of the scale, and neither rater used
1 or 2. Expert A gave $\{5{:}72, 4{:}8\}$ (90\% fives, SD 0.30) and Expert B
$\{5{:}52, 4{:}25, 3{:}3\}$ (65\% fives, SD 0.56). The raters agreed exactly on
52/80 samples (65.0\%) and within one point on 77/80 (96.3\%), yet the
chance-corrected coefficients are near zero: Cohen's weighted $\kappa = 0.0719$ and
Krippendorff $\alpha_{\text{interval}} = -0.0359$. This is the high-agreement,
low-$\kappa$ paradox~\citep{feinstein1990paradox}: when almost all ratings fall in one
category, expected chance agreement is itself high, so $\kappa$ and $\alpha$ collapse
even though the raters rarely differ by more than one point. Gwet's AC1 for the
five-versus-below-five decision is 0.46, and the linearly weighted AC2 on the 1--5
scale is 0.87~\citep{gwet2008ac1}. The disagreement is also directional: in 24 of the
28 non-identical pairs, Expert A gave 5 and Expert B gave 4 or 3, a difference in
severity rather than random noise. We therefore read the ratings as showing that both
teachers found nearly all candidates realistic, not as a reliable ranking among them.

\paragraph{What determined the selected set}
After flag caps, 41 of the 80 candidates received 5 from both raters and were tied at
the top of the ranking. The diversity constraints excluded none of them, so the 25
selected seeds are the first 25 of these 41 in generation order. With the tie broken
at random instead (1{,}000 draws), the same rule reproduces on average only 15.5 of
the 25 seeds (95\% interval 13--19). Generation order also shaped domain coverage.
The selected set covers four of the six domains (Communication 11, Motor 8,
Cognitive 3, Social-Emotional 3) and contains no Sensory Processing or Self-Care seed,
although the tied pool held four of each. Under random tie-breaking, all six domains
appear in 91.8\% of draws, and no draw covers four or fewer.

The ratings did shift the selection relative to rating-free rules. Drawing 25
flag-free candidates at random yields on average 16.6 seeds rated 5 by both experts
(versus 25) and overlaps the actual set by 10.4 seeds. Because the scale was
compressed, however, the corresponding difference in Expert B's mean score is small
(5.0 versus 4.68), and archetype entropy is essentially unchanged (2.84 versus 2.82).
Expert B's ratings alone reproduce 23 of the 25 seeds and Expert A's alone 15, so in
practice the selection reflects the stricter rater. The remaining rules from
Section~\ref{sec:phaseB-method} confirm the role of tie order: ranking by the mean score
first reproduces all 25 seeds and dropping the flag caps 21, but reversing the tie order
reproduces only 9 (while covering five domains). Ignoring the ratings entirely and drawing 25 of the 80
eligible candidates at random overlaps the actual set by 7.9 seeds on average.

\subsection{Phase C: Expert Gold Parsing and Feature Diffusion}
Phase C collected both experts' final annotations, producing 25 entries of
\texttt{parsed\_gold.json} via \texttt{06\_parse\_annotations.py}.
Field-level agreement between the two experts was 100\% (25/25) on primary domain,
concern count, and secondary-goal presence, and no sample raised a validation warning,
so the adjudication queue stayed empty; Expert~A's annotations serve as gold and
Expert~B's as an audit copy. The original 15/10 train/hold-out split is retained.
Diffusion (Section~\ref{sec:phaseC-method}) then produced 567 valid examples out of
585, giving a training set of 582 examples.

\subsection{AI-Assisted Annotation Burden}
\label{sec:annot}
To avoid mistaking LLM annotations for gold labels, GPT-5.5 is used solely as a
\emph{weak pre-annotator}. All 25/25 AI-assisted annotation packages were completed
and compared field-level with expert-final annotations via exact match:
overall \textbf{acceptance rate 64.25\%}, correction rate 35.75\%.
Annotation time was collected over 50 expert $\times$ sample observations
(\texttt{time\_coverage\_rate}=1.0). The blank-labeling baseline is 30.0 min/case;
AI-assisted time averaged 2.846 min, yielding a \textbf{time reduction rate of 90.5\%}.
This analysis does not claim AI drafts equal expert quality; it quantifies the degree
to which AI converts expert labor from blank-page labeling to adjudication.

\begin{figure*}[!t]
\centering
\includegraphics[width=0.85\textwidth]{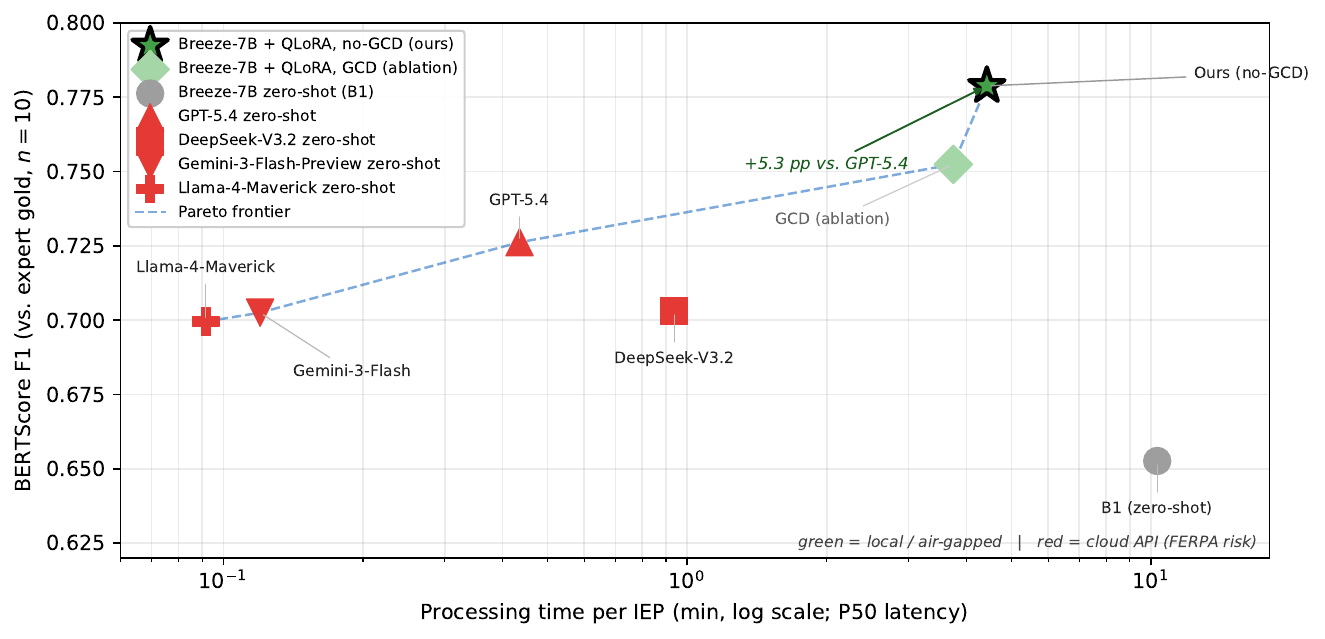}
\caption{Time--quality trade-off: per-IEP processing time (P50 latency, minutes, log
scale) vs.\ BERTScore F1 against expert gold ($n=10$). Green markers: local /
air-gapped deployment; red: cloud API (FERPA risk). The Pareto frontier runs
Llama-4-Maverick $\rightarrow$ Gemini-3-Flash-Preview $\rightarrow$ GPT-5.4
$\rightarrow$ Breeze-7B + QLoRA with GCD $\rightarrow$ without GCD ($\bigstar$, ours).
The fine-tuned local model reaches the highest F1 while staying air-gapped, at the cost
of the longest latency on the frontier.}
\label{fig:pareto-cq}
\end{figure*}

\subsection{Cross-Vendor Baselines}
Table~\ref{tab:baseline} reports schema compliance, fill rate (the mean share of
required schema fields that are non-empty), and BERTScore F1~\citep{zhang2020bertscore}
on the hold-out set ($n=10$), whose interviews are synthetic like the training data.
BERTScore F1 is computed with the \texttt{bert-base-chinese} model (library-default
layer, no baseline rescaling) on three fields --- the extracted \emph{primary concern raw quote},
the generated \emph{annual goal text}, and the generated \emph{semester goals}, where
each predicted semester goal is greedily matched to its best-scoring gold goal --- and
averaged. Cloud models are run zero-shot with the same instruction as the local models.
Ladder consistency and LLM-judge scores are left to future work. We report score
differences in points, where one point is 0.01 F1.

\begin{table}[!t]
\caption{Cross-Vendor Baseline Comparison on the Hold-out Set ($n=10$).
BERTScore F1 raw, un-rescaled (\texttt{bert-base-chinese}); within-column comparable only.}
\centering
\footnotesize
\setlength{\tabcolsep}{2.5pt}
\begin{tabular}{@{}lccc@{}}
\toprule
\textbf{Model} & \textbf{Schema} & \textbf{Fill} & \textbf{BERTScore} \\
& \textbf{Compliance} & \textbf{Rate} & \textbf{F1} \\
\midrule
\multicolumn{4}{@{}l}{\emph{Cloud APIs, zero-shot}} \\
GPT-5.4                  & 100\%          & 100\%          & 0.726 \\
DeepSeek-V3.2            & 100\%          & 100\%          & 0.703 \\
Gemini-3-Flash-Preview   & 90\%$^\dagger$ & 90\%$^\dagger$ & 0.703 \\
Llama-4-Maverick         &  90\%          &  90\%          & 0.700 \\
\midrule
\multicolumn{4}{@{}l}{\emph{Local, Breeze-7B}} \\
zero-shot, GCD (B1)      & 100\%          & n/a            & 0.653 \\
+ QLoRA, GCD             & 100\%$^\ast$   & n/a            & 0.752 \\
\textbf{+ QLoRA, no-GCD} & \textbf{100\%} & n/a            & \textbf{0.779} \\
\bottomrule
\end{tabular}
\label{tab:baseline}
\end{table}

\begin{figure*}[!t]
\centering
\includegraphics[width=\textwidth]{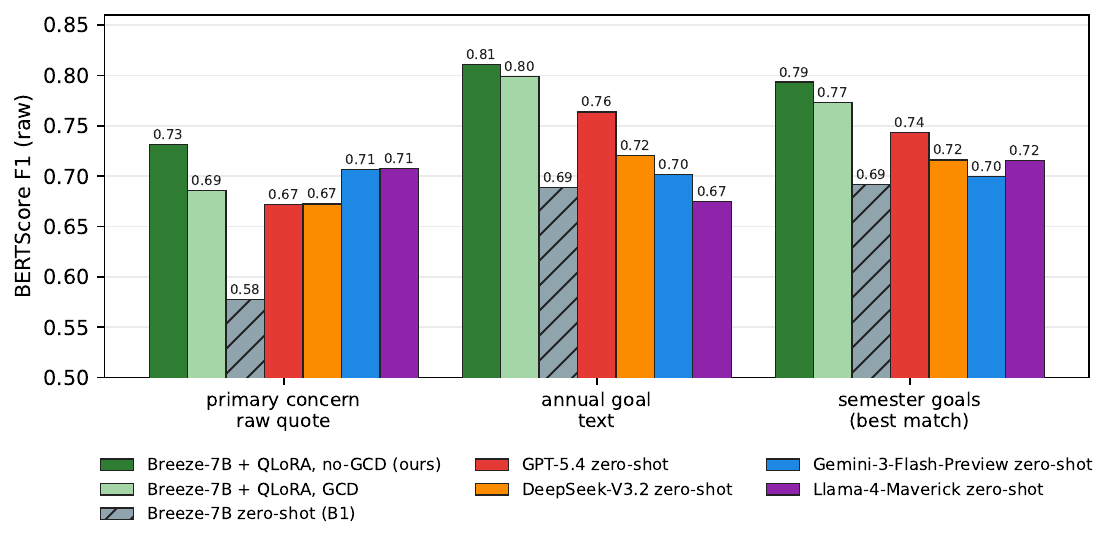}
\caption{Per-field BERTScore F1 across seven configurations ($n=10$ hold-out set,
\texttt{bert-base-chinese}, raw, no rescale): \emph{primary concern raw quote},
\emph{annual goal text}, and \emph{semester goals (best-match)}.
Breeze-7B + QLoRA without GCD scores highest on all three fields. With GCD it still
leads on the two generative fields (annual and semester), while Llama-4-Maverick and
Gemini-3-Flash-Preview are marginally ahead on the extractive concern quote field.}
\label{fig:field-fill}
\end{figure*}

\noindent\emph{Notes:} GPT-5.4 and DeepSeek-V3.2 pass the schema check on 10/10 outputs;
Llama-4-Maverick 9/10 (one failure: multiple concerns simultaneously marked as primary);
Gemini-3-Flash-Preview$^\dagger$ 9/10 (one failure: \texttt{secondary\_ladder} $<$ 2 semester goals;
evaluated with \texttt{thinking\_budget=0} on Vertex~AI to avoid MAX\_TOKENS budget
exhaustion from implicit reasoning tokens).
$^\ast$GCD 92.7\% on the $n=55$ stress set (Section~\ref{sec:stress}); both fine-tuned
paths pass 10/10 on the hold-out set.
Fill rate ``n/a'' for local models reflects enforced structural completeness, not
explicit field-content verification. Fig.~\ref{fig:field-fill} decomposes per-field F1.

\paragraph{Uncertainty.}
Because the hold-out set has only ten records, Table~\ref{tab:ci} tests every
comparison with a paired bootstrap confidence interval and an exact sign-flip
permutation test over records. The no-GCD path (\textbf{0.779}) is ahead of all four
zero-shot cloud baselines, and every one of these margins has a CI that excludes zero; the narrowest,
over GPT-5.4, is $+0.053$ (95\% CI $+0.012$ to $+0.098$, $p = 0.047$, ahead on 7 of 10
records). The GCD path (0.752) is ahead of DeepSeek-V3.2, Gemini-3-Flash-Preview, and
Llama-4-Maverick, but its 2.6-point margin over GPT-5.4 is not resolved ($p = 0.096$),
and neither is the 2.6-point gap between the two fine-tuned paths ($p = 0.061$). The
fine-tuning gain under GCD ($+0.100$ over B1, ahead on 10 of 10 records) is the
clearest effect.

\begin{table}[!t]
\caption{Paired comparisons of overall BERTScore F1 on the hold-out set ($n=10$;
$n=9$ for Gemini-3-Flash-Preview and Llama-4-Maverick, whose schema-invalid output is
not scored). $\Delta$: mean paired difference (first system minus second); CI: 95\% percentile bootstrap over
records (10{,}000 resamples); $p$: exact two-sided sign-flip permutation test; Wins:
records on which the first system scores higher. Not corrected for multiple
comparisons.}
\centering
\footnotesize
\setlength{\tabcolsep}{2pt}
\begin{tabular}{@{}lrcrr@{}}
\toprule
\textbf{Comparison} & $\Delta$ & \textbf{95\% CI} & $p$ & \textbf{Wins} \\
\midrule
\multicolumn{5}{@{}l}{\emph{QLoRA, no-GCD minus}} \\
GPT-5.4 & 0.053 & [0.012, 0.098] & 0.047 & 7/10 \\
DeepSeek-V3.2 & 0.076 & [0.035, 0.119] & 0.012 & 8/10 \\
Gemini-3-Flash-Preview & 0.071 & [0.026, 0.124] & 0.012 & 8/9 \\
Llama-4-Maverick & 0.082 & [0.046, 0.124] & 0.004 & 9/9 \\
GCD & 0.026 & [0.004, 0.057] & 0.061 & 8/10 \\
\midrule
\multicolumn{5}{@{}l}{\emph{QLoRA, GCD minus}} \\
GPT-5.4 & 0.026 & [0.000, 0.057] & 0.096 & 7/10 \\
DeepSeek-V3.2 & 0.049 & [0.023, 0.081] & 0.008 & 9/10 \\
Gemini-3-Flash-Preview & 0.048 & [0.016, 0.088] & 0.020 & 8/9 \\
Llama-4-Maverick & 0.056 & [0.026, 0.096] & 0.004 & 9/9 \\
B1 (no adapter) & 0.100 & [0.079, 0.123] & 0.002 & 10/10 \\
\bottomrule
\end{tabular}
\label{tab:ci}
\end{table}

\paragraph{Stress set.}
\label{sec:stress}
Besides the hold-out set, both fine-tuned paths are run on a stress set of 55 transcripts:
the Phase~A candidates that were not selected as seeds ($80 - 25$). They come from the
same GPT-5.4 generator and Kimi-K2.6 pre-screen as the seeds, were never used for
training or the hold-out set, and have no expert annotations, so on this set we measure only
schema validity, latency, and verbatim quoting. By transcript length, 3 are short
($<$1{,}500 characters), 47 mid-length, and 5 long ($\geq$2{,}500); 7 are Sensory
Processing and 8 Self-Care cases.

\subsection{Quote Faithfulness}
\label{sec:quote}
Each extracted concern carries a \texttt{raw\_quote} that is meant to be the parent's
own words. We check this automatically: a quote counts as \emph{verbatim} if it is a
substring of the input transcript after Unicode NFKC normalization and whitespace
removal. The expert annotations meet this for all 17 quotes in the hold-out set (and all
42 in the 25 annotated seeds), so verbatim quoting is the convention the training
data teaches.

\begin{table}[!t]
\caption{Share of predicted parent quotes found verbatim in the input transcript
(hold-out set, $n=10$). $^\dagger$Nine of the ten outputs contained concern quotes.}
\centering
\footnotesize
\begin{tabular}{@{}lrr@{}}
\toprule
\textbf{Model} & \textbf{Verbatim} & \textbf{Rate} \\
\midrule
Expert gold                              & 17/17 & 100.0\% \\
\midrule
\multicolumn{3}{@{}l}{\emph{Local, Breeze-7B}} \\
zero-shot, GCD (B1)                      &  7/28 & 25.0\% \\
+ QLoRA, GCD                             & 17/21 & 81.0\% \\
\textbf{+ QLoRA, no-GCD}                 & \textbf{21/24} & \textbf{87.5\%} \\
\midrule
\multicolumn{3}{@{}l}{\emph{Cloud APIs, zero-shot}} \\
GPT-5.4                                  & 15/32 & 46.9\% \\
DeepSeek-V3.2                            & 38/40 & 95.0\% \\
Gemini-3-Flash-Preview$^\dagger$         & 13/32 & 40.6\% \\
Llama-4-Maverick$^\dagger$               & 34/36 & 94.4\% \\
\bottomrule
\end{tabular}
\label{tab:quote}
\end{table}

Table~\ref{tab:quote} shows that, with GCD in both arms, fine-tuning raises the verbatim
rate of the same base model from 25.0\% to 81.0\%; the no-GCD path reaches 87.5\%. On
the stress set the fine-tuned model stays at 89.2\% (107/120) without GCD and 87.4\% (90/103) with it.
The fine-tuned model does not lead on this measure: DeepSeek-V3.2 and
Llama-4-Maverick quote verbatim more often (95.0\% and 94.4\%), while GPT-5.4 and
Gemini-3-Flash-Preview do so less often (46.9\% and 40.6\%; 59.4\% each when
punctuation differences are also ignored). All systems receive the same instruction,
which asks for concerns drawn from the parent's own words but does not state that the
quote must be copied exactly, so the zero-shot rates reflect how each model reads that
instruction. The measure checks only that a quote exists in the transcript, not that
it is the most relevant one.

\subsection{Ablation Studies}
\label{sec:ablation}

To test the core design choices of CGFD, we ran four ablations, split into
\emph{data-side} (A1/A2, on diffused examples) and \emph{model-side} (B1/B2, on the
hold-out and stress sets):

\begin{itemize}[leftmargin=*]
  \item \textbf{A1: Few-shot exemplar count.}
        \emph{exemplars=0} vs.\ \emph{exemplars=5}: tests the contribution of
        in-context learning (ICL) to style transfer.
  \item \textbf{A2: FeatureProfile injection.}
        \emph{profile=ON} vs.\ \emph{profile=OFF}: tests the contribution of
        corpus-grounded statistical constraints to output authenticity.
  \item \textbf{B1: QLoRA fine-tuning.}
        \emph{adapter=ON} vs.\ \emph{adapter=OFF (zero-shot)}: tests the contribution
        of domain fine-tuning to schema compliance and semantic similarity; both arms
        use GCD.
  \item \textbf{B2: Grammar-Constrained Decoding.}
        \emph{GCD=ON} vs.\ \emph{GCD=OFF (free-form)}: tests the marginal contribution
        of constrained decoding to schema compliance and operational reliability.
\end{itemize}

A1 and A2 share the full diffusion run's GPT-5.4 generator, domain quota, and random seed; B1 and
B2 share its adapter, prompt, and inputs. In each case only the controlled variable
changes (\emph{ceteris paribus}).

\begin{table*}[!t]
\caption{Ablation results. Data side: A1/A2 ($n=200$) vs.\ the full diffusion run
         (Main, $n=567$). Model side: B1/B2 vs.\ the fine-tuned GCD path (Main) on the
         hold-out set ($n=10$) and the stress set ($n=55$). n/a: not applicable;
         ---: not run.}
\centering
\footnotesize
\begin{tabular}{@{}lccccc@{}}
\toprule
\textbf{Metric}
   & \textbf{Main}
   & \textbf{A1}
   & \textbf{A2}
   & \textbf{B1}
   & \textbf{B2} \\
   & \emph{full; GCD}
   & \emph{ex=0}
   & \emph{no prof.}
   & \emph{no QLoRA; GCD}
   & \emph{no GCD} \\
\midrule
$n$ records              & 567   & 200   & 200   & 10   & 10 \\
schema compliance        & 100\% & 100\% & 100\% & 100\% & 100\% \\
Distinct-2               & 0.072 & 0.146 & 0.126 & n/a  & n/a \\
Distinct-4               & 0.497 & 0.627 & 0.624 & n/a  & n/a \\
Self-BLEU(4-gram, $n=50$)& 0.613 & 0.595 & 0.606 & n/a  & n/a \\
prompt-leak rate         & 0.4\% & 0.5\% & 0.0\% & n/a  & n/a \\
neg.-behavior rate       & 0.0\% & 0.0\% & 0.0\% & n/a  & n/a \\
primary domain entropy   & 2.343 & 2.357 & 2.357 & n/a  & n/a \\
\midrule
latency P50 (sec., $n=10$)& 224.8 & n/a  & n/a   & 618.5 & 265.5 \\
latency P95 (sec., $n=10$)& 345.7 & n/a  & n/a   & 829.9 & 438.7 \\
schema rate ($n=55$ stress)& 92.7\% & n/a & n/a  & ---  & \textbf{100\%} \\
P50 ($n=55$ stress, sec.) & 117.2 & n/a  & n/a   & ---  & \textbf{77.3} \\
P95 ($n=55$ stress, sec.) & 343.6 & n/a  & n/a   & ---  & \textbf{94.3} \\
BERTScore F1 ($n=10$)     & 0.752 & n/a  & n/a   & 0.653 & \textbf{0.779} \\
\bottomrule
\end{tabular}
\label{tab:ablation}
\end{table*}

\noindent
\textbf{A1/A2: Data-side observations.}
Three patterns emerge across the data-side ablations:
(i)~schema compliance holds at 100\% in all three configurations;
(ii)~Self-BLEU differs by at most 1.8 points (0.595--0.613); Distinct-N is lower for
Main than for A1/A2 (Distinct-4 0.497 vs.\ 0.627/0.624), largely because Main's corpus is
$2.8\times$ larger, which lowers the share of unique $n$-grams, while A1 and A2 differ by
at most 2.0 points;
(iii)~prompt-leak and negative-behavior rates are near zero.
This does not negate the value of in-context exemplars and FeatureProfile injection;
rather, it constrains their observable contribution level: current data-side metrics
can only support the claim that \emph{schema validity is primarily guaranteed by
built-in prompt rules and Pydantic validation, with no observed necessity of
ICL/profile for schema compliance}. Whether exemplars and profile improve expert
style fidelity requires additional metrics (length distribution, mastery patterns, or
human style ratings). This contribution decoupling offers a methodological insight for
generalizing CGFD to other low-resource domains: schema enforcement and style anchoring
are independent and can be calibrated separately per application.

\noindent
\textbf{B1 --- QLoRA contribution to content fidelity.}
All three local configurations achieve 10/10 schema compliance on the hold-out set, so
on this set schema compliance does not depend on QLoRA, and B2 shows that it does not
require GCD either. BERTScore F1 separates the adapter's content contribution: with GCD in both arms, the fine-tuned model scores \textbf{0.752} and B1
(no adapter) \textbf{0.653}, a 9.9-point gap that holds on all ten records
(Table~\ref{tab:ci}). On the same metric, the GCD path is ahead of DeepSeek-V3.2,
Gemini-3-Flash-Preview, and Llama-4-Maverick and level with GPT-5.4 at $n=10$, while the
no-GCD path is ahead of all four. QLoRA thus brings the local 7B model's content
alignment to the level of zero-shot cloud models, and above it on the no-GCD path;
whether this reflects style fidelity in particular is not measured
(Section~\ref{sec:limits}).

\noindent
\textbf{B2 --- GCD did not help this adapter.}
B2 (no GCD) achieves \textbf{55/55 = 100\%} schema compliance on the stress set,
with P50 = 77.3~s and P95 = 94.3~s. The GCD path yields 51/55 = 92.7\%, with
P50 = 117.2~s and P95 = 343.6~s. All four failures are on the GCD side; an exact
McNemar test on the paired inputs gives $p = 0.125$, so at this sample size the
reliability difference is suggestive rather than statistically significant. On this
set the no-GCD path is \textbf{34\% faster at the median}, its 95th-percentile latency
is \textbf{$3.6\times$ lower}, and its latency band is tighter (P95/P50 ratio 1.22
vs.\ 2.93). The latency advantage does not carry over to the hold-out set, where
no-GCD is slower at the median (265.5~s vs.\ 224.8~s; Table~\ref{tab:ablation}).

All four GCD failures share one failure mode: the JSON output is cut off by the
\texttt{max\_new\_tokens=2048} budget before the ladder closes (\texttt{EOF while
parsing} at character 7{,}485--7{,}725). Because both paths share the tokenizer and
the budget, the truncations do not reflect a higher token cost of Chinese text; the
saved outputs show instead that GCD changed what the model generated. The no-GCD outputs never exceed 589 tokens (median 453, about
2.2 characters per token), and for the four inputs on which GCD failed they are
392--493 tokens, about 1{,}000 characters, long. Under GCD the same inputs produced
output that was still unfinished after 2{,}048 tokens and about 7{,}500 characters.
Even re-serialized compactly, the 51 successful GCD outputs have a median of 642
tokens against 453 for no-GCD on the same inputs, so GCD lengthens the output in
general and, on four inputs, beyond the budget.

Two further observations bear on the cause without isolating it. First, the failures
are not stable across runs. In later reruns of the four failing inputs, three produced
valid outputs of 470--707 tokens under the same 2048-token budget, and with a 3072-token
budget all four were valid while using at most 683 tokens, so a larger budget is not
what resolved them; these reruns are not guaranteed to reproduce the original decoding
pipeline exactly. Second, the adapter was trained with a 2048-token cap that cuts off the
end of most training targets (Section~\ref{sec:phaseD}), so it rarely saw a complete
output end. In follow-up runs, an adapter trained with the same recipe at a 4096-token
cap, which truncates no example, produced 55/55 valid outputs under GCD, with a
GCD-to-unconstrained length ratio of 0.99, against 1.42 for the 2048-cap adapter. The
same follow-up work found that the GCD outcome for the 2048-cap adapter also changes
with how the decoding pipeline compiles the schema, such as the order of its
properties. Training truncation and pipeline details are therefore both candidate
causes; with one training run per setting and run-to-run variation, this paper cannot
separate them. We therefore recommend the no-GCD path for deployment until the effect is
understood (Section~\ref{sec:discussion}).

\subsection{Pareto Failure Analysis}
\begin{figure}[!t]
\centering
\includegraphics[width=\columnwidth]{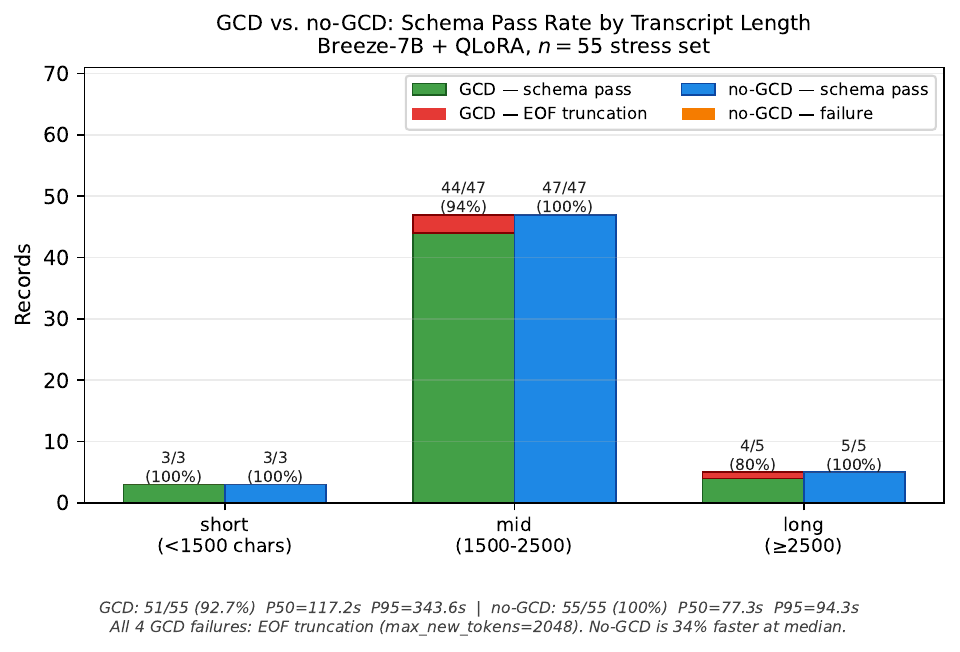}
\caption{GCD vs.\ no-GCD pass/fail counts on the $n=55$ stress set
by transcript character length (short $<$1500 / mid 1500--2500 / long $\geq$2500).
The no-GCD path passes 55/55 across all bins; GCD fails 4 inputs (3 mid, 1 long),
all via the same EOF token-budget truncation. The footer gives P50/P95 latency; at the
median, no-GCD is 34\% faster (77.3~s vs.\ 117.2~s). The four failures are truncations
at the token budget, not content errors.}
\label{fig:pareto-failure}
\end{figure}

To characterize the operational envelope, we bin the stress-set transcripts by
character length --- short ($<$1500), mid (1500--2500),
long ($\geq$2500) --- and compare pass/fail counts between the GCD and no-GCD paths.
Fig.~\ref{fig:pareto-failure} shows that the no-GCD path passes all 55 inputs across
all length bins, while the GCD path fails 4/55 (three mid-length and one long input).
All four failures share the EOF truncation mode described in Section~\ref{sec:ablation}.
Raising \texttt{max\_new\_tokens} is not the remedy: at a 3072-token budget, later
reruns completed all four inputs within 683 tokens, well inside the original budget
(Section~\ref{sec:ablation}).

\subsection{Proposed Self-Preference Bias Audit}
\label{sec:spb}
This paper does not use an LLM judge; this subsection specifies how a future
judge-based evaluation should be audited. If GPT-5.4 were used as LLM-as-Judge, it might favor output
from its own distribution~\citep{wataoka2024,chen2025dbg}. Following the DBG Score of
\citet{chen2025dbg}, the protocol uses dual-expert gold scores as quality proxies. For
judge model $J$ evaluating the response $r_{X,i}$ of model $X$:
\begin{align}
\text{DBG}_J(X)        &= \mathbb{E}_i\!\left[ J(r_{X,i}) - g_i \right] \label{eq:dbg}\\
\text{centered-DBG}_J  &= \text{DBG}_J(J) \notag\\
                       &\quad - \mathbb{E}_{X' \neq J}\!\left[\text{DBG}_J(X')\right] \label{eq:cdbg}
\end{align}
where $g_i$ is a dual-expert quality score for output $i$, still to be collected (the
realism ratings of Section~\ref{sec:phaseB} rate transcripts, not model outputs).
$\text{DBG}_J(J)$ quantifies $J$'s bias toward its own output;
$\text{centered-DBG}_J$ further subtracts $J$'s overall scoring tendency toward other
models, separating self-preference from general strictness. Full SPB quantification
requires cross-model judge/gold score tables (future work).

\section{Discussion and Limitations}
\label{sec:discussion}

\subsection{Stylistic Mode Collapse: An Empirical Observation}
During development, we observed that including a concrete example utterance in the
\emph{system prompt} (e.g., ``Teacher, what should we do about this'') caused GPT-5.4
to treat it as a mandatory template, producing that phrase in over 25 of 60 samples.
This is consistent with the \emph{typicality bias} described in Verbalized
Sampling~\citep{vs2025}: even with 8 archetypes and diversified prompts, a literal
example sentence in the prompt becomes a strong prior that overrides the diversity
quotas. The six-layer detection mechanism reduces such phrases to below 10\% of the
final batch, but eliminating them requires removing literal examples from the prompt.
This has practical implications for any research that uses LLMs for batch synthesis.

\subsection{GCD Did Not Help This Adapter}
Grammar-constrained decoding was added as a reliability guardrail, yet on the stress set
it produced the only four invalid outputs (51/55 vs.\ 55/55; exact McNemar
$p = 0.125$). The failures are truncations, and Section~\ref{sec:ablation} shows that
they arise because GCD lengthened what the model wrote, not because Chinese text costs
more tokens: a complete no-GCD output needs only about 450 tokens. Follow-up runs point
to two candidate causes, the 2048-token training cap and details of the decoding
pipeline, which this paper cannot separate. On the hold-out set
the two paths are close. The no-GCD path scores 2.6 points higher in BERTScore F1, a gap
that $n=10$ does not resolve (Table~\ref{tab:ci}), and its median latency is higher
(265.5~s vs.\ 224.8~s). Taken together, \textbf{for this adapter at the current token
budget, the no-GCD path was more reliable on the stress set and no worse in semantic
similarity; its latency advantage holds on the stress set only}.

For practitioners, the lesson is narrower than a CJK-specific token penalty: constrained
decoding can change how much a fine-tuned model writes, and whether it does may depend on
how the model was trained (here, with most training targets cut off at 2048 tokens) and
on how the decoding pipeline compiles the schema. Training without truncation and
comparing GCD output lengths with unconstrained ones are cheap safeguards. We tested one
adapter per setting; separating these factors with repeated training runs is the main
prerequisite before GCD can be recommended for this task.

The finding parallels our prior \textsc{ARID/NASD} work~\citep{authors2026arid}, where
a designed-in component also degraded performance in an unanticipated deployment
context. In both cases the degradation was invisible on a small, clean test set and
appeared only on a larger stress set, which makes stress-set evaluation a reusable
discipline rather than a one-off check.

\subsection{Privacy Engineering Value for Deployable AI}
Many teachers already use AI for IEP work~\citep{cdt2025}; when that AI runs in the
cloud, student data is exposed to FERPA/IDEA risk. Our design --- local Breeze-7B +
QLoRA fine-tuning with offline inference --- keeps raw interview data inside the school
network. The adapter file \texttt{adapter\_model.safetensors} is approximately 160~MB,
and the full training directory, including checkpoints and optimizer state, is
approximately 498~MB. Training and all local inference ran on a single RTX 4070 (12~GB)
workstation, with a median latency of 224.8~s (GCD) to 265.5~s (no-GCD) per interview on
the hold-out set. Air-gapped operation is therefore feasible on one consumer GPU;
deployment on smaller edge hardware, such as the Jetson-class devices used
in~\citet{authors2026arid}, has not been tested.

\subsection{Methodological Generalizability}
This work shares its methodological backbone with our prior
\textsc{ARID/NASD}~\citep{authors2026arid} (corpus-grounded statistical extraction
$\rightarrow$ LLM-constrained diffusion $\rightarrow$ edge fine-tuning and deployment),
while the target domain, schema structure, evaluation metrics, and noise sources differ
entirely. The transfer suggests that CGFD offers an actionable route for educational
service systems. Domain-agnostic evidence still requires multi-institution data,
content-quality judges, and cross-vendor baselines before the approach can serve as a
general backbone for \emph{information-processing workflow automation} under an
industrial engineering paradigm.

\subsection{Limitations}
\label{sec:limits}

\subsubsection{Sample Scale}
Dual-expert rating and annotation cover 25 seeds: 15 training seeds and the 10-interview
hold-out set. Twenty-five expert-annotated interviews are too few for a comprehensive
benchmark of Traditional Chinese special-education NLP, and the ten-record hold-out set
limits how finely systems can be separated (Table~\ref{tab:ci}). Future work will expand
to $\geq 50$ interviews rated with a calibrated rubric (Section~\ref{sec:lim-rating}).

\subsubsection{Rating Scale Compression and Seed Selection}
\label{sec:lim-rating}
The realism ratings used for seed selection were compressed at the top of the scale
(Section~\ref{sec:phaseB}). Raw agreement was high (96.3\% within one point), but
chance-corrected agreement was near zero, and the ratings could not rank the 41
candidates that both teachers scored 5 after flag caps; generation order decided among
them. Two consequences follow. First, the $\tau$ curve documents how the seeds were
selected, but it is not evidence that they are the most realistic candidates. Second,
the 15 training seeds and the 10-interview hold-out set contain no Sensory Processing or
Self-Care case. These two domains enter training only through diffused examples (22 and
40 of the 567) and evaluation only through the stress set (7 and 8 interviews). The
ratings also measure transcript realism, not IEP quality. Future rounds will
(i)~anchor each scale point with example transcripts, including deliberately weak ones,
and hold a calibration round before independent rating, with boundary cases adjudicated
following LEAP~\citep{lee2023common}; (ii)~break selection ties at random within domain
strata; and (iii)~add a separate expert rating of generated IEP quality.

\subsubsection{Annotation Anchoring}
The gold annotations started from GPT-5.5 drafts: the experts accepted 64.25\% of draft
fields unchanged and corrected the rest (Section~\ref{sec:annot}). The gold annotations
may therefore lean toward the drafts' wording, which could favor systems whose outputs
resemble GPT-5.5's style, including the fine-tuned model trained on these annotations
and possibly GPT-5.4. Annotating a subset from scratch would measure the size of this
effect.

\subsubsection{Single Institution and Synthetic Data}
All interviews in this study --- training seeds, diffused examples, the hold-out set, and
the stress set --- are synthetic and come from the same GPT-5.4 generation pipeline, so
the evaluation measures performance on in-distribution synthetic interviews;
performance on real parent--teacher interviews is untested. The expert annotations come
from a single educational institution and may reflect institution-specific writing
conventions. Validation on real, multi-institution data requires sustained effort on
IRB procedures and data-sharing agreements.

\subsubsection{Ablation Sample Scale and Lexical Metric Boundaries}
A1/A2 use $n=200$ while the full diffusion run uses $n=567$, so Distinct-N comparisons
are confounded by corpus size. Self-BLEU is more robust as a sample-level metric; its
differences are at most 1.8 points, but no uncertainty interval was computed. The
defensible implication of the data-side ablations is that schema validity does not
depend on ICL or the FeatureProfile; style fidelity still lacks direct measurement.

\subsubsection{Training Truncation}
The 2048-token training cap truncated 444 of the 524 training examples, so the adapter
learned from targets whose final part was cut off (Section~\ref{sec:phaseD}). All
results in this paper come from this adapter; an adapter trained without truncation may
differ in quality as well as in its behavior under GCD.

\subsubsection{Prompt-Leak Residual}
Despite six-layer detection and hard/soft regeneration, some high-frequency phrases
learned by GPT-5.4 appear in variant forms in the final batch. Complete resolution may
require logit-level intervention during diffusion, such as $n$-gram priors injected at
decoding time~\citep{ngram2026}; this is left to future work.

\subsubsection{Clinical Decision Boundary}
The system is positioned as a tool that reduces teachers' documentation burden. Any
final IEP must be reviewed, revised, and signed off by a special-education teacher; the
system is not intended to replace professional judgment, consistent with the
human-in-the-loop direction recommended by the CDT report~\citep{cdt2025}.

\section{Conclusion}

This work builds a low-resource pipeline that drafts IEP goals from Traditional Chinese
parent--teacher interviews. Centered on Corpus-Grounded Feature Diffusion (CGFD), it
transfers the NASD methodology from industrial maintenance work
orders~\citep{authors2026arid} to an educational service setting. Its four contributions
match those listed in the introduction: (i)~a six-stage pipeline from seed
overproduction and dual-expert $\tau$ selection to local QLoRA fine-tuning, with a
traceable record of how the seeds were selected; (ii)~$8 \times 5 \times 12$
three-layer quotas with six-layer detection and hard/soft regeneration to counter
stylistic mode collapse, reported through Distinct-4 and Self-BLEU; (iii)~GPT-5.5
pre-annotation with expert adjudication, in which experts accepted 64.25\% of draft
fields and corrected 35.75\%, with a time reduction of 90.5\%; and (iv)~a multi-vendor
evaluation design with a specified \emph{centered}-DBG protocol for auditing LLM-as-Judge
self-preference bias, to be run once judge scores are collected.

Three findings stand out. First, on the hold-out set the fine-tuned no-GCD path
(BERTScore F1 0.779) is ahead of all four zero-shot cloud baselines (GPT-5.4 0.726,
DeepSeek-V3.2 0.703, Gemini-3-Flash-Preview 0.703, Llama-4-Maverick 0.700), with paired
bootstrap CIs that exclude zero; the narrowest margin, over GPT-5.4, is $+0.053$ (95\% CI
$+0.012$ to $+0.098$). Second, the gain comes from fine-tuning: under the same GCD
decoding, F1 rises from 0.653 to 0.752 and the share of verbatim parent quotes from
25.0\% to 81.0\% (87.5\% without GCD); the GCD path is ahead of three cloud baselines and
level with GPT-5.4, and the 2.6-point gap between the two fine-tuned paths is not
resolved at $n=10$. Third, GCD did not help this adapter: on the stress set it produced
the only four invalid outputs (92.7\% vs.\ 100\% schema compliance, P50 117.2~s vs.\
77.3~s; exact McNemar $p = 0.125$), all truncations at the 2048-token budget on inputs
for which the no-GCD output needed fewer than 500 tokens. The failures did not recur
reliably in later reruns, and an adapter trained at a 4096-token cap showed no such loss,
which points to the 2048-token training cap and the decoding pipeline as candidate
causes. As in our prior \textsc{ARID}
work~\citep{authors2026arid}, only the stress set exposed this deployment constraint.

The system runs on a single consumer-GPU workstation, keeping interview data off cloud
APIs and reducing FERPA/IDEA privacy risk. These results rest on synthetic interviews, a
ten-record hold-out set, and BERTScore, which measures similarity to expert annotations
rather than the educational quality of the goals (Section~\ref{sec:limits}). Within
those bounds, the work fills a gap in Traditional Chinese special-education NLP and
offers a reusable backbone that integrates \emph{data quality control, system
robustness, and deployability} under an industrial engineering paradigm.

Future work: (i)~expand seeds to multi-institution annotation, re-rated with an
anchored, calibrated rubric and selected with domain-stratified random tie-breaking;
(ii)~separate the effects of the training cap and the decoding pipeline on GCD
reliability with repeated training runs; (iii)~report bootstrap 95\% CIs on the
expanded hold-out set; (iv)~introduce logit-level diversity intervention during
diffusion~\citep{ngram2026}; (v)~conduct a semester-length field deployment study with
special-education teachers.

\section*{Acknowledgments}
The authors thank the two anonymous special-education experts for their independent
scoring contributions during seed annotation and $\tau$ selection, and MediaTek
Research for open-sourcing the Breeze-7B model weights.

\bibliography{refs}

\end{document}